\documentclass{gretsi}
  \usepackage{enumitem}
   \usepackage[english,french]{babel}   %
  \usepackage{times}			%
 \usepackage{todonotes}
\usepackage{multirow}
\usepackage{booktabs}
\usepackage{amsmath}
\usepackage{stmaryrd}
\usepackage{amsfonts}
\usepackage{hyperref}
\usepackage{comment}
  \usepackage[maxbibnames=3, maxcitenames=3, style=numeric, backend=biber, bibencoding=utf8]{biblatex}
\usepackage{setspace}

\usepackage{bm}
\usepackage{dsfont}

\def\skA{{\mathcal{A}}}

\def\O{{\mathcal{O}}}
\def\Cx{{\mathbb{C}}}

\def\empX{{\mathcal{\hat X}}}

\def\P{{\mathcal{P}}}

\newcommand{\intint}[1]{\left \llbracket#1\right \rrbracket}

\def\rmDelta{{\mathbf{\Delta}}}

\def\1{\bm{1}}

\def\rva{{\mathbf{a}}}

\def\rvc{{\mathbf{c}}}

\def\rvh{{\mathbf{h}}}
\def\rvu{{\mathbf{i}}}

\def\rvs{{\mathbf{s}}}

\def\rvu{{\mathbf{u}}}

\def\rvw{{\mathbf{w}}}
\def\rvx{{\mathbf{x}}}

\def\rvz{{\mathbf{z}}}

\def\rmA{{\mathbf{A}}}

\def\rmH{{\mathbf{H}}}

\def\rmN{{\mathbf{N}}}

\def\rmU{{\mathbf{U}}}

\def\rmW{{\mathbf{W}}}

\def\rmTheta{{\mathbf{\Theta}}}

\def\valpha{{\bm{\alpha}}}

\DeclareMathAlphabet{\mathsfit}{\encodingdefault}{\sfdefault}{m}{sl}
\SetMathAlphabet{\mathsfit}{bold}{\encodingdefault}{\sfdefault}{bx}{n}

\newcommand{\E}{\mathbb{E}}

\newcommand{\R}{\mathbb{R}}

\title{Compressive Clustering with an Optical Processing Unit}

\author{\coord{Luc}{Giffon}{1},
        \coord{R\'{e}mi}{Gribonval}{1}}

\address{\affil{1}{Univ Lyon, Inria, CNRS, ENS de Lyon, UCB Lyon 1,
LIP UMR 5668, F-69342, Lyon, France}}

\email{luc.giffon@ens-lyon.fr, remi.gribonval@ens-lyon.fr}

\englishabstract{ Compressive clustering is a technique used to find centroids in a dataset by first computing a low-dimensional sketch summarizing the data then learning only from the sketch. The sketch can be computed using the celebrated random Fourier features method. In this paper, we explore the use of Optical Processing Units (OPU) to compute the random Fourier features for sketching, and the adaptation of the overall pipeline to this setting. We also propose some tools to help tuning a critical hyper-parameter of compressive clustering. 
}

\englishabstract{ 
We explore the use of Optical Processing Units (OPU) to compute random Fourier features for sketching, and adapt the overall compressive clustering pipeline to this setting. We also propose some tools to help tuning a critical hyper-parameter of compressive clustering. 
}

\frenchabstract{Nous proposons une proc\'edure de \textit{sketching} pour le \textit{clustering} compressif utilisant des processeurs optiques (OPU) pour calculer les composantes de Fourier aléatoires. Cette proc\'edure fait intervenir une strat\'egie novatrice qui permet de choisir efficacement l'\'echelle du sketch.}

\begin{document}

\maketitle
\section{Introduction}
\label{sec:introduction}

\vspace{-0.3cm}

A popular tool to address high-dimensional machine learning problems is the use of Random Fourier Features (RFF), which provide nonlinear feature maps exploiting random projections %
 \cite{rahimi2007random}.  This paper focuses on the use of RFF to build sketches for compressive $K$-means clustering \cite{keriven2018sketching}. 
 
 A sketch is the expected value of a feature map $\Phi(\cdot)$ on samples of a data distribution. Clusters can be learnt from a sketch built using properly chosen RFF  \cite{keriven2018sketching}, and  \textit{sketching} and \textit{learning from the sketch} can happen on different machines. This gives rise \cite{schellekens2021extending} to an asynchronous framework in which the (possibly private \cite{chatalic2021compressive}) sketch is constructed on a low-resource \textit{sketching device} and sent to a central server where the clusters can be recovered from it. Yet, sketching for asynchronous compressive clustering is currently not a turnkey procedure. 

A first major obstacle is to choose the distribution of the random matrix involved in RFF. Despite existing heuristics \cite{chatalic2020efficient,keriven2018sketching}, tuning this distribution is still something of an art %
essentially boiling down to trying several distributions for the random matrix, to build as many sketches and learn from them before picking the best. This procedure is expensive, and all the more unsatisfying as it contradicts the very philosophy of sketched learning which is to make only one pass on the dataset.

The second issue with sketching is its reliance on a very large matrix to compute RFF in high dimension. Chatalic et al. in  \cite{chatalic2018large} alleviate this burden for compressive clustering by re-using a classical method from the kernel community, based on replacing a dense random Gaussian matrix with products of structured and sparse random matrices associated to fast linear transforms. Here we investigate an alternate approache exploiting a novel co-processor called ”Optical Processing Unit (OPU)”, which is dedicated to perform random matrix multiplications by leveraging the physical properties of light traveling through a scattering medium. In theory, the OPUs random multiplications have a linear time and space complexity with respect to the sum of the input and output dimensions \cite{ohana2020kernel}.
An OPU also has a constant power consumption of 30W. 

OPUs can speed up RFF for kernel machines \cite{ohana2020kernel} and it seems natural to apply them to compressive clustering. However, %
each OPU is associated to its unique draw of a random matrix, so it is not possible to have the same OPU on the sketching device and on the central server. More importantly, some steps of the main algorithms to learn from a sketch require explicit access to the underlying random matrix, while an OPU only implements multiplication by this matrix.
Fortunately, the matrix associated to an OPU can be estimated via calibration, which leads us to propose a new asynchronous compressive clustering pipeline. Its main steps, and accordingly the contributions of this paper, are the following  (see also Figure~\ref{fig:sequence}):
\begin{figure}
\includegraphics[width=0.47\textwidth]{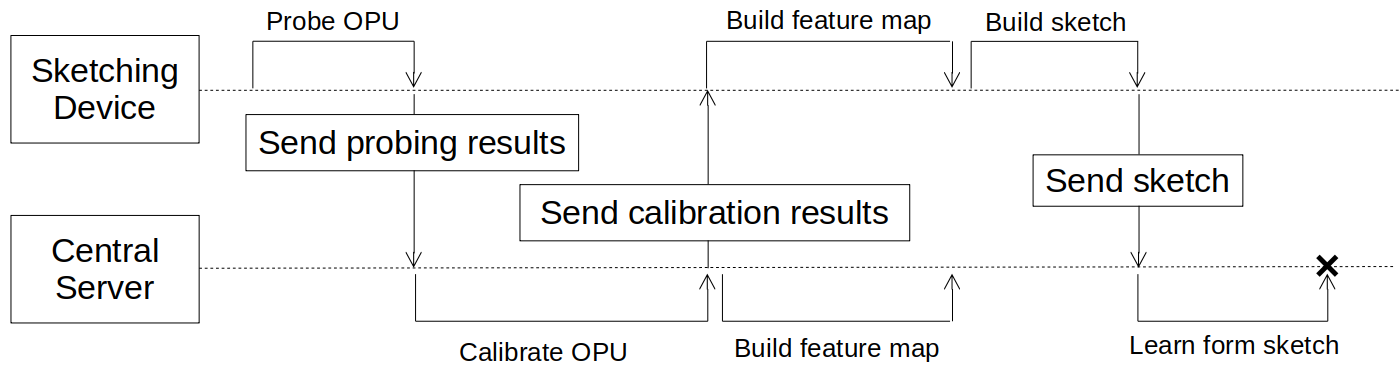}
\caption{Sequence diagram of the proposed pipeline. }
\label{fig:sequence}
\end{figure}
\begin{enumerate}[wide, labelwidth=!,labelindent=1pt, noitemsep,topsep=1pt,parsep=0pt,partopsep=0pt]
\item Feature maps are built on the sketching device and the central server by leveraging calibration (Section~\ref{sec:build_fm});
\item We propose a trick to simultaneously compute sketches for different RFF distributions on the sketching device (Section~\ref{sec:sigma_grid_search});
\item We propose a criterion to choose the best distribution and send the corresponding sketch to the central server (Section~\ref{sec:entropy});
\item The server learns the clusters from the sketch (Experimental results in Section~\ref{sec:results}).
\end{enumerate}
\vspace{-0.4cm}
\subsection{Sketching for compressive clustering}
A training collection made of $N$ observations in $D$ dimensions  $\{\rvx_i\}_{i=1}^{N}$ is represented by its empirical data distribution $\P_{\empX} := \frac{1}{N}\sum_{i=1}^{N}\delta_{\rvx_i}$. 
The RFF map $\Phi: \R^D \rightarrow \Cx^M$ is defined\footnote{The exponential $\exp: \Cx \rightarrow \Cx$  is applied entry-wise on its input.} as 
$\Phi(\rvx) := \exp(-i\rmW\rvx)$,
where $\rmW \in \R^{M \times D}$ is a random matrix yet to be specified. The sketch $\hat\rvz \in \Cx^M$ is then defined as the empirical average of this feature map over the training set:
\vspace{-0.2cm}
\begin{equation}
\label{eq:sketching}
\hat\rvz := \frac{1}{N} \sum_{i=1}^{N} \Phi(\rvx_i) = \E_{\rvx \sim \P_{\empX}}[\Phi(\rvx)].
\end{equation}

For compressive $K$-means clustering, $M$ is usually in the order of $\O(KD)$, with $K$ the prescribed number of clusters. Each line $\rvw_j$ of $\rmW$ (so called \textit{frequencies} due to the link with Fourier analysis) is typically a unit-norm direction $\rvu_j \in \R^D$ sampled uniformly and independently on the unit-sphere, multiplied by a radius $\rho_j \in \R^+$ sampled from a distribution akin to a folded Gaussian \cite{keriven2018sketching} with variance 1,  and finally rescaled by $\frac{1}{\sigma}$. Stacked together, these frequencies give the following definition for the matrix $\rmW$:
 \vspace{-0.1cm}

\begin{equation}
\label{eq:W}
\mathbf{W} := \frac{1}{\sigma}\rmDelta\rmU, 
\end{equation}

\vspace{-0.1cm}
\noindent with $\rmDelta := \mathtt{diag}(\rho_1, \ldots, \rho_M)$, $\rmU := \left[ \rvu_1, \ldots, \rvu_M \right]^T \in \R^{M\times D}$. The distribution of $\rmW$ is parameterized by the value of the sketching scale $\sigma$, which must be tuned with respect to the dataset.

\vspace{-0.4cm}
\subsection{Learning centroids from a RFF sketch}
\vspace{-0.2cm}

\textbf{$K$-means clustering} seeks $K$ centroids $\rmTheta := \{\rvc_1, \ldots, \rvc_K\}$ as close as possible, in the sense of the Euclidean distance, to the observed data points. This corresponds to minimizing
\vspace{-0.1cm}

\begin{equation}
\label{eq:risk_kmeans}
\mathcal{\hat R}(\rmTheta) := \sum_{i=1}^{N} \|\rvx_i - \rmTheta(\rvx_i)\|^2
\end{equation}

\noindent where $ \rmTheta(\cdot)$ maps its input to the closest centroid. 
Alternatively, $K$-means clustering can be  interpreted as minimizing the Wasserstein-2 distance   between the empirical data distribution $\P_{\empX}$  and a mixture of $K$ Diracs \cite{vayer2021controlling} $\P_{\rmTheta, \valpha} := \sum_{k=1}^{K}\alpha_k \delta_{\rvc_k}$ with $\sum_{k=1}^{K}\alpha_k = 1$. 
Similarly in {\bf compressive $K$-means clustering}, the centroids are chosen by minimizing the distance $ \left\| \skA(\P_{\rmTheta, \valpha}) - \skA(\P_{\empX}) \right\|_2^2$ between the \textit{sketches} of these distributions.
\noindent where the \emph{sketching operator} $\skA : \P \rightarrow \Cx^M$ gives a vector representation, the sketch, of its input distribution. The data sketch~\eqref{eq:sketching} is $\hat\rvz = \skA(\P_{\empX})$, while $\skA(\P_{\rmTheta, \valpha}) = \sum_{k=1}^{K}\alpha_k \Phi(\rvc_k)$.

Compressive Learning Orthogonal Matching Pursuit with Replacement (CLOMP-R) \cite{keriven2018sketching} is a heuristic algorithm to minimize $ \left\| \skA(\P_{\rmTheta, \valpha}) - \skA(\P_{\empX}) \right\|_2^2$. Given a feature map $\Phi(\cdot)$, the only input to CLOMP-R is the empirical sketch $\hat\rvz$ (and \emph{not} the training dataset). CLOMP-R involves gradient descent exploiting the derivative of the feature map $\Phi(\cdot)$. This is a challenging requirement when implementing sketches with an OPU.

\vspace{-0.4cm}

\subsection{Optical Processing Units}
\label{subsec:opu}
\vspace{-0.2cm}

An "\textbf{Optical Processing Unit (OPU)}" \cite{saade2016random} is a hardware implementing a function $A(\cdot)$ to approximately perform a random matrix product  with an input vector $\rvx \in \R^D$. It leverages the diffusion properties of light traveling through a scattering medium, and can be modeled as:
\vspace{-0.2cm}

\begin{equation}
\label{eq:OPU}
A(\rvx) = D(\rmA E(\rvx) + \boldsymbol{\epsilon}).
\end{equation}

\noindent Functions $E(\cdot)$ and $D(\cdot)$ are respectively bit plane encoding and decoding functions needed because the physical, low level, OPU only takes as input binary vectors (vectors in $\{0,1\}^D$); vector $\boldsymbol{\epsilon} \in \R^M$ is a  Gaussian noise sampled for each input $\rvx$ and caused by the analogic implementation of the function $A(\cdot)$; finally, and more importantly, the \emph{transmission matrix} $\rmA \in \R^{M \times D}$ 
is sampled from a centered Gaussian distribution, and can be considered as fixed once the OPU is built. 

\vspace{-0.5cm} 
\paragraph{Calibration} An estimate $\hat{A}$ of the transmission matrix $\rmA$ can be obtained via calibration: similarly to \cite{gupta2020fast}, the linearity of the OPU can be leveraged to create a set of linear equations allowing to recover the coefficients in $\rmA$. This is done by applying the transmission matrix $\rmA$ to the columns of the Hadamard matrix $\rmH$ in dimension $D$ (up to zero padding to the first greater power of $2$) \cite{kanjilal1995adaptive}  and solving the $M$ resulting linear equations.

\vspace{-0.4cm}
\section{Building OPU-based feature maps}
\label{sec:build_fm}
\vspace{-0.2cm}

In the proposed pipeline (see~Figure~\ref{fig:sequence}), the sketching device benefits the low-power property of the OPU to sketch the data with an energy-efficient feature map $\Phi_{\mathtt{OPU}}(\cdot)$. The central server, which has no resource constraint, builds a sort of {\em digital twin feature map} $\hat \Phi_{\mathtt{OPU}}(\cdot) \approx \Phi_{\mathtt{OPU}}(\cdot)$ to learn from the sketch.

\vspace{-0.4cm}
\paragraph{The sketching device feature map $\Phi_{\mathtt{OPU}}(\cdot)$}

The core idea of using the OPU for sketching is to take advantage of the OPU function $A(\rvx)$ to perform the random matrix multiplication in $\Phi(\rvx)$ in place of the $\rmW\rvx$ product. However, the transmission matrix $\rmA$ of the OPU has random Gaussian entries while $\rmW$ is different, see~\eqref{eq:W}. Hence, the output of  $A(\cdot)$ must be rescaled by a diagonal matrix in order to fit the definition of $\rmW$. %

Denote $\hat\rva_{i}$ the rows of the calibrated transmission matrix $\hat\rmA$ (see below) and let
 $\rmN$ be the diagonal with entries 
$\rmN_{i,i} := \frac{1}{\|\hat\rva_i\|_2}$.
We propose to re-scale the output of $A(\cdot)$ with $\rmN$ such that $\rmN A (\rvx)$ can be identified with
 $\rmU \rvx$. Scaling again with %
$\rmDelta$ and $\frac{1}{\sigma}$ to mimic~\eqref{eq:W} gives the following expression for $\Phi_{\mathtt{OPU}}$:

\begin{equation}
\label{eq:phiopu}
\Phi_{\mathtt{OPU}}(\rvx) := \exp\left(-i\frac{1}{\sigma}%
\rmDelta
\rmN A(\rvx)\right).
\end{equation}
\vspace{-0.6cm}
\paragraph{The calibration step}
To determine $\rmN$, the central server estimates $\hat \rmA$ (see Section~\ref{subsec:opu}) by exploiting probed vectors 
sent from the sketching device to the server. Such vectors are obtained from the application of the OPU to the columns of the Hadamard matrix $\rmH$. From the calibrated matrix, the server can send back the (diagonal of) the matrix $\rmDelta\rmN$ to the sketching device, as it is needed to compute $\Phi_{\mathtt{OPU}}(\cdot)$ according to~\eqref{eq:phiopu} and to compute a sketch $\hat\rvz_{\mathtt{OPU}} := \frac{1}{N} \sum_{i=1}^{N} \Phi_{\mathtt{OPU}}(\rvx_i)$ of the empirical data distribution like in Equation~\eqref{eq:sketching}. In the process, the sketching device never stores the calibrated transmission matrix $\hat\rmA$ nor anything requiring more than $\O(M)$ bytes of memory. 

\vspace{-0.4cm}
\paragraph{The digital twin feature map $\hat \Phi_{\mathtt{OPU}}(\cdot)$} 
The matrix $\rmDelta\rmN$ is also used on the server to define a mirroring feature map\linebreak $\hat\Phi_{\mathtt{OPU}}(\rvx):=\exp\left(-i%
\frac{1}{\sigma}\rmDelta\rmN
\hat\rmA\rvx\right)$.
This feature map has a known derivative so it can be used to run CLOMP-R and learn the centroids from the sketch $\hat\rvz_{\mathtt{OPU}}$ received from the sketching device.

\vspace{-0.5cm}
\section{Choosing the sketching scale}
\vspace{-0.2cm}
\label{sec:sigma}
To achieve good clustering performance one needs to tune the scale $\sigma$ used for the random projection matrix~ \eqref{eq:W} \cite{chatalic2021compressive, gribonval2021sketching, keriven2018sketching}.  In this section, we propose a two-fold procedure to efficiently and accurately select a good scale $\sigma$ for compressive clustering.

\begin{figure}[t!]
\begin{center}
\includegraphics[width=0.45\textwidth]{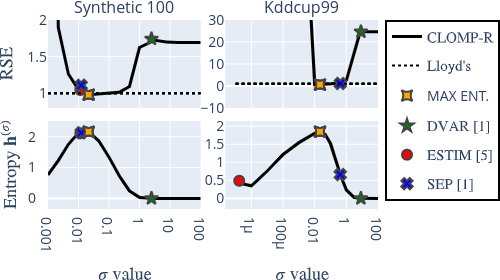}
\vspace{-0.2cm}
\caption{Entropy of sketch VS clustering quality on a synthetic dataset (see Section~\ref{sec:experimental_details}) and  KDDCUP99 dataset \cite{tavallaee2009detailed}.}
\label{fig:entropy}
\end{center}
\end{figure}

\vspace{-0.4cm}
\subsection{Efficient grid-search for sketching scale }
\label{sec:sigma_grid_search}

An experiment on a single dataset needs a total of $S$ different sketching scales $\sigma_{1}, \ldots, \sigma_{S}$ (Equation~\eqref{eq:W}) corresponding to build as many feature maps $\{\Phi^{(1)}, \ldots, \Phi^{(S)}\}$ and to produce as many sketches. Computing these sketches can be prohibitively expensive if done naively one sketch after another. Fortunately, the bulk of the computation can be shared between the $S$ feature maps and producing the sketches can be much more efficient if they are all computed simultaneously. 
With the method we propose, the total complexity (in %
 flops) of computing these $S$ sketches on silicium (CPU or GPU) drops 
from $NS \O(MD)$ to $N({\O(SM)}+\O(MD))$. %
For large $D$, this divides the complexity by $S$. %
This speed is enabled by the fact that for each observation $\rvx$ in the dataset
the product ${\rmDelta}\rmU \rvx$ can be computed only once 
 (with a cost $\O(MD)$),  then its Kronecker product with the vector $\rvs :=\left[\frac{1}{\sigma^{(1)}}, \ldots, \frac{1}{\sigma^{(S)}}\right]^T$ is computed in $SM$ flops, and the pointwise exponential $\exp(-i\rvs \otimes (\rmDelta\rmU \rvx))$ returns all RFF feature maps at once with $\O(SM)$ flops ($\otimes$ denotes the Kronecker product).
This yields the claimed gain in flops,
with a small overhead in space complexity of $MS$ (instead of $M$) corresponding to storing in memory all the $S$ sketches at once, each of dimension $M$. 

Does this trick also reduce the time cost of computing $S$ sketches with an OPU ? Suppose that the time cost of applying $A(\cdot)$ is equivalent to $C_{\mathtt{OPU}} (M+D)$ flops and that $M \gg D$ (in compressive $K$-means \cite{chatalic2020efficient} $M=100KD$ is typical). One can check that the trick multiplies the time cost by $1/C_{\mathtt{OPU}}+1/S$. This leads to a ``win-win'' situation: if the OPU is moderately efficient --associated to somewhat large constant $C_{\mathtt{OPU}} \geq S$-- then the proposed grid-search still pays off;  otherwise $C_{\mathtt{OPU}} \ll S$ leads to a very fast implementation of $\Phi_{\mathtt{OPU}}$ even though grid-search cannot benefit from further acceleration.

In practice, we observed a $\times 2$ speed up on GPU when $S=10$, $D=200$ and $M=500D$. Using a real OPU, we evaluated $C_{\mathtt{OPU}} = (\mathtt{Time~of} A(\rvx)) / \left((\mathtt{Time~of~one~flop}) \times (M + D)\right) \approx 6000$. This big  constant $C_{\mathtt{OPU}}$ led to an observed $\times12$ speed up in the same setting as for the GPU. This was of crucial importance to run the experiments of Section~\ref{sec:results}.

\vspace{-0.4cm}
\subsection{Entropy as a proxy for sketch quality}
\label{sec:entropy}

Sketching an empirical distribution~\eqref{eq:sketching} is equivalent to sampling its characteristic function on the locations defined by the frequencies in $\rmW$  (see e.g. \cite{keriven2018sketching}). In the high and low frequency regimes (large / small norm $\|\rvw_{j}\|_{2}$), the characteristic function has values close to 0 or 1 respectively. Since this holds whatever the distribution \cite{keriven2018sketching}, sampling the characteristic function of any distribution with too high or too low frequencies does not allow to distinguish between distributions.

It is thus natural to assume that a well chosen scale $\sigma$ should ensure that the magnitude of the sketch coefficients are well distributed between 0 and 1. %
To promote this property, we propose to use an entropy measure.
Concretely, the interval $[0, 1]$ is split into $B$ non overlapping bins of size $\frac{1}{B}$. Then, the sketch coefficients are assigned a bin with respect to their magnitude. Hence, a sketch coefficient defines a discrete random variable $\mathcal{B} \in \intint{B}$. From one sketch obtained with a particular $\sigma$, we can make an estimation of the probability $P^{(\sigma)}_b$ of any sketch coefficient to fall inside the $b^{th}$ bin and then apply the classical Shanon formula for entropy estimation \cite{shannon2001mathematical},
$\rvh^{(\sigma)}(\mathcal{B}) = - \sum_{b=1}^{B}P^{(\sigma)}_b \log\left(P^{(\sigma)}_b\right)$.
We expect this entropy to be smallest when $\sigma$ leads to magnitudes either all close to 0 or all close to 1. Vice-versa, the highest entropy should lead to a ``best'' $\sigma$.

On Figure 3, the top plots display the Relative Squared Error (RSE) (see Metrics in Section~\ref{sec:experimental_details}) of CLOMP-R with respect to $\sigma$. We observe that the range of $\sigma$ for which the performance of CLOMP-R is equivalent to the Lloyd algorithm (RSE close to one) is relatively narrow. From the observation of the bottom plot --displaying the entropy of the sketch with respect to $\sigma$-- it appears clearly that the best performance is indeed reached when the entropy of the sketch is at its peak (MAX ENT. marker). Comparing results on both dataset, we remark that other existing methods to select $\sigma$ give unpredictable performance depending on the datasets: DVAR is the global data variance \cite{chatalic2020efficient} and doesn't perform good on any dataset; ESTIM is an iterative method proposed in \cite{keriven2018sketching} and perform well on the synthetic dataset but very bad on KDDCUP99; SEP is an expectation of the separation between clusters\cite{chatalic2020efficient} and perform relatively well in both dataset but still worse than the entropy based estimator.

\vspace{-0.6cm}
\section{Assessing the quality of centroids}
\label{sec:results}
\label{sec:experimental_details}

\vspace{-0.3cm}

The proposed pipeline involving the OPU and its calibration is evaluated by comparing the quality of the obtained centroids compared with that of the vanilla CLOMP-R on two synthetic Gaussian mixture datasets. Before proceeding to the actual results, we provide some experimental details.
\vspace{-0.6cm}
\paragraph{Data} We use a synthetic data generation routine implemented in the  \texttt{pycle} toolbox \cite{pyclegithub}.
It samples observations from a mixture of $K=5$ isotropic Gaussians in dimension $D=10$ or $100$ with a
inter-cluster/intra-cluster variance ratio equal to 10.

 \vspace{-0.6cm}
\paragraph{Code} We adapted \texttt{pycle} \cite{pyclegithub} to support \texttt{pytorch} \cite{pytorch} and \texttt{lightonml} \cite{lightonmlgithub} . Our version is available online \cite{pyclegithubluc}.

\vspace{-0.6cm}
\paragraph{Hyper-parameters} We choose the scale $\sigma$ among a log-range of $S=10$ possible values in $[10^{-3}, 1]$  by applying the entropy criterion (Section~\ref{sec:entropy}). We repeat the experiments for $10$ samples of %
the radius matrix $\rmDelta$ from the distribution proposed in \cite{keriven2018sketching}. Centroids in the CLOMP-R algorithm as well as those of the Lloyd algorithm are initialized randomly in the hyper-cube containing the dataset. We use a sketch size $M=100KD$.

\vspace{-0.6cm}
\paragraph{Metrics \& Competing methods} We use three metrics computed with respect to centroids returned by the Lloyd's algorithm which serve as an ideal clustering method: (i) the Relative Squared Error (RSE) e.g. the ratio $\frac{\mathcal{\hat R}(\rmTheta_{\mathtt{OPU}})} {\mathcal{\hat R}(\rmTheta_{\mathtt{Lloyd}})}$ \cite{chatalic2020efficient} (see Equation~\eqref{eq:risk_kmeans}), the Adjusted Mutual Information(AMI) \cite{vinh2010information} and the Wasserstein distance (W-Dist.)\cite{kantorovich1960mathematical}. 
Three variants of the pipeline are compared: CLOMP-M (``Matrix''), where a GPU is used to sketch by computing RFF and the vanilla CLOMP-R algorithm is used to learn from the resulting sketch; CLOMP-SO (Simulated OPU) using a simulated OPU that should behave like an ideal OPU, however with no noise; CLOMP-RO (Real OPU) using an actual OPU to compute the sketches. 

 We compare the results with 
 two baselines that randomly sample : (i) each centroid uniformly from all observations in the dataset (RAND-DATA); or (ii) the $k$-th centroid uniformly from all observations {\em from the $k$-th cluster} (RAND-CLS; this highly informed baseline is feasible since the clusters have been synthetically generated). 
These controls
serve as a reference points to assess the quality of clustering with CLOMP-R.

\vspace{-0.6cm}
\paragraph{Results}

\begin{table}[tb!]
\begin{center}
\resizebox{0.48\textwidth}{!}{\begin{tabular}{c|ccc|ccc}
\toprule
\multirow{2}{*}{\textbf{Method}}   & \multicolumn{3}{c|}{Synthetic 10} & \multicolumn{3}{c}{Synthetic 100} \\ 
\cmidrule{2-7} 
&  \textbf{RSE}&\textbf{AMI}&\textbf{W-Dist.}&\textbf{RSE}&\textbf{AMI}&\textbf{W-Dist.} \\
\midrule
CLOMP-M & 1.02($\pm$ 0.0) & 0.84($\pm$ 0.1) & 0.03($\pm$ 0.1) & 0.98($\pm$ 0.1) & 0.96($\pm$ 0.1) & 0.07($\pm$ 0.1) \\ 
CLOMP-SO & 1.02($\pm$ 0.0) & 0.84($\pm$ 0.1) & 0.03($\pm$ 0.1) & 0.98($\pm$ 0.1) & 0.96($\pm$ 0.1) & 0.08($\pm$ 0.1) \\ 
CLOMP-RO & 1.04($\pm$ 0.0) & 0.80($\pm$ 0.1) & 0.04($\pm$ 0.0) & 1.24($\pm$ 0.1) & 0.85($\pm$ 0.1) & 0.17($\pm$ 0.1) \\ 
\midrule
RAND-CLS & 1.53($\pm$ 0.1) & 0.43($\pm$ 0.1) & 0.22($\pm$ 0.1) & 1.88($\pm$ 0.2) & 0.88($\pm$ 0.1) & 0.31($\pm$ 0.1) \\ 
RAND-DATA & 1.69($\pm$ 0.2) & 0.37($\pm$ 0.1) & 0.25($\pm$ 0.1) & 2.13($\pm$ 0.3) & 0.63($\pm$ 0.1) & 0.39($\pm$ 0.1) \\ 
\bottomrule
\end{tabular}}
\end{center}
\vspace{-0.4cm}
\caption{Results: Averages and standard deviations.}
\label{tab:results}
\end{table}

Table~\ref{tab:results} shows that, on the Synthetic 10 dataset,\linebreak CLOMP-RO has results similar to the classical CLOMP-M method. On the Synthetic 100 dataset however, while  CLOMP-SO behaves similarly to CLOMP-M, the performance of CLOMP-RO is degraded suggesting that it suffers from the analogic noise of the real OPU \eqref{eq:OPU}. Yet, the performance of \linebreak CLOMP-RO remains noticeably better than the baseline RAND-CLS. This confirms that \textit{some} clustering information is caught by the CLOMP-RO pipeline as RAND-CLS is highly informed, contrary to RAND-DATA whose results are clearly worse.

\vspace{-0.6cm}
\section{Conclusion and perspectives}
\vspace{-0.2cm}

We propose a compressive clustering pipeline exploiting the energy-efficiency of OPUs to sketch and using the power of CPU/GPU to learn from sketches. %
While current OPU hardware is sized to fit in a server rack, the obtained results prompt for the development of a minified and embeddable OPU to be deployed on a low-resource sketching device. Further challenges include the investigation and possible correction of the performance degradation  observed in higher dimension, as well as the validation of the approach on diverse real world datasets.

\vspace{-0.5cm}
\paragraph{Acknowledgement} AllegroAssai ANR project ANR-19-CHIA-0009 supported this work. This work was granted access to the pilot project between LightOn and IDRIS / GENCI enabling us to use a LightOn OPU at the Jean Zay supercomputer. We thank the support of the Centre Blaise Pascal's IT test platform at ENS de Lyon which operates the SIDUS solution \cite{quemener2013sidus}.

\vspace{-0.3cm}
\begin{spacing}{0.8}
\printbibliography
\end{spacing}

\end{document}